\documentclass[10pt,twocolumn,letterpaper]{article}

\usepackage[pagenumbers]{wacv}
\usepackage[T1]{fontenc}
\usepackage{xcolor}  
\providecommand{\tcam}{{\normalfont\textsc{TCAM}}\xspace}
\providecommand{\car}{{\normalfont\textsc{CAR}}\xspace}

\definecolor{figBlue}{RGB}{122,158,196}    
\definecolor{figAmber}{RGB}{196,154,114}   
\definecolor{figGreen}{RGB}{122,173,136}   
\definecolor{figViolet}{RGB}{158,140,186}  
\definecolor{figSlate}{RGB}{136,152,170}   
\newcommand{\cBlue}[1]{\textcolor{figBlue}{\textbf{#1}}}
\newcommand{\cAmber}[1]{\textcolor{figAmber}{\textbf{#1}}}
\newcommand{\cGreen}[1]{\textcolor{figGreen}{\textbf{#1}}}
\newcommand{\cViolet}[1]{\textcolor{figViolet}{\textbf{#1}}}

\definecolor{wacvblue}{rgb}{0.21,0.49,0.74}
\usepackage[pagebackref,breaklinks,colorlinks,allcolors=wacvblue]{hyperref}

\usepackage{graphicx}
\usepackage{booktabs}
\usepackage{multirow}
\usepackage{pifont}
\usepackage{wrapfig}
\usepackage{amsmath}
\usepackage{amssymb}
\usepackage{xspace}
\usepackage{fontawesome5}
\newcommand{\codelink}{%
  \href{https://github.com/ostadabbas/TCAM-Track-and-Caption-Any-Motion}%
       {\faGithub\;\texttt{TCAM-Track-and-Caption-Any-Motion}}%
}

\newcommand{\cmark}{\ding{51}}
\newcommand{\xmark}{\ding{55}}

\def\wacvPaperID{3674}
\def\confName{WACV}
\def\confYear{2027}

\title{Track and Caption Any Motion:
Open-Vocabulary Spatiotemporal Captioning
via Trajectory-Conditioned Generation}

\author{%
  Bishoy Galoaa\\
  Northeastern University\\
  {\tt\small galoaa.b@northeastern.edu}
  \and
  Sarah Ostadabbas\thanks{Corresponding author.}\\
  Northeastern University\\
  {\tt\small s.ostadabbas@northeastern.edu}
}

\begin{document}

\newcommand{\figLandscape}{%
\begin{figure}[t]
  \centering
  \includegraphics[width=0.88\columnwidth]{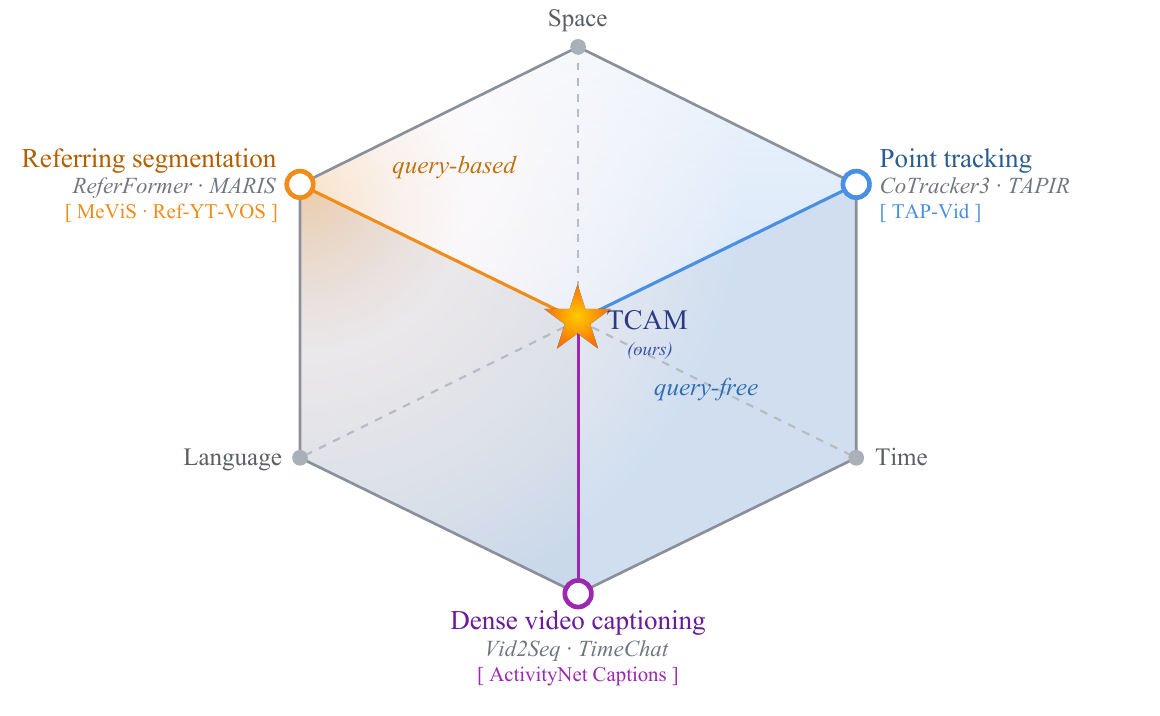}
  \caption{%
    The video-understanding landscape. Space, time, and language are the three
    axes a complete motion description must span, and prior work occupies the
    two-axis corners: point trackers
    (\cBlue{CoTracker3}~\cite{karaev2025cotracker3},
    \cBlue{TAPIR}~\cite{doersch2023tapir}) localize without language; referring
    segmentation (\cAmber{ReferFormer}~\cite{wu2022referformer},
    \cAmber{MARIS}~\cite{wang2024maris}) grounds from a query; dense captioners
    (\cViolet{Vid2Seq}~\cite{yang2023vid2seq},
    \cViolet{TimeChat}~\cite{ren2024timechat}) describe from clip-level
    features. \tcam\ reaches the remaining corner, evaluated on the benchmark
    native to each axis.%
  }
  \label{fig:landscape}
\end{figure}%
}

\newcommand{\figPipeline}{%
\begin{figure*}[t]
  \centering
  \includegraphics[width=\textwidth]{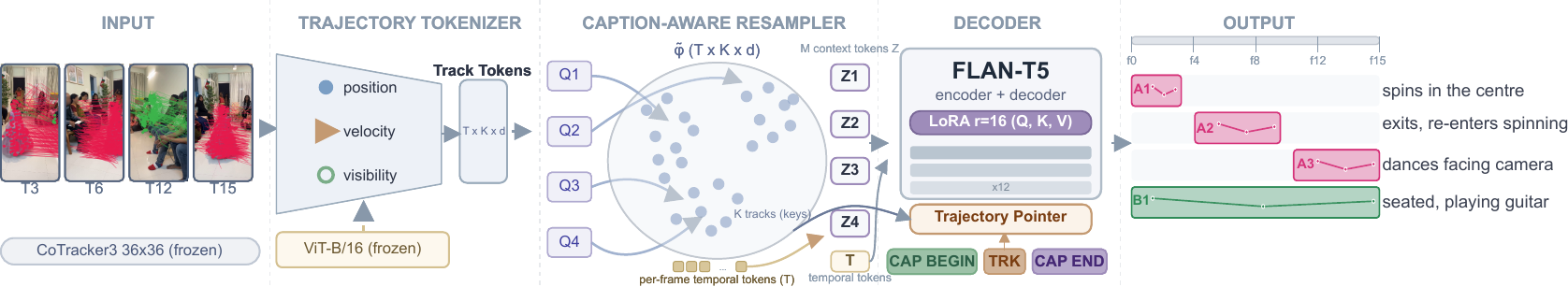}
  \caption{%
    \tcam\ pipeline.
    Frozen CoTracker3 ($K{=}1296$ tracks) and ViT-B/16 feed the
    \textit{VisionMotionEncoder}, producing trajectory tokens $\tilde{\Phi}$
    (position, velocity, visibility). \car\ compresses $\tilde{\Phi}$ into $M$
    context tokens $\mathbf{Z}$, which join per-frame temporal tokens at the
    FLAN-T5 encoder input; the decoder (LoRA, $r{=}16$) emits captions
    interleaved with \cGreen{\texttt{CAP\_BEGIN}}, \cAmber{\texttt{TRK}},
    \cViolet{\texttt{CAP\_END}} and time tokens \texttt{<T$_i$>}.
    Each \texttt{TRK} fires a pointer head over the same $K$ track tokens,
    yielding $\hat{\mathbf{p}}\in\mathbb{R}^{K}$.
    \textit{Output:} A1--A3 trace one subject across non-adjacent intervals;
    B1 is a concurrent second subject.
  }
  \label{fig:pipeline}
\end{figure*}%
}

\newcommand{\figResampler}{%
\begin{figure}[t]
  \centering
  \includegraphics[width=0.8\columnwidth]{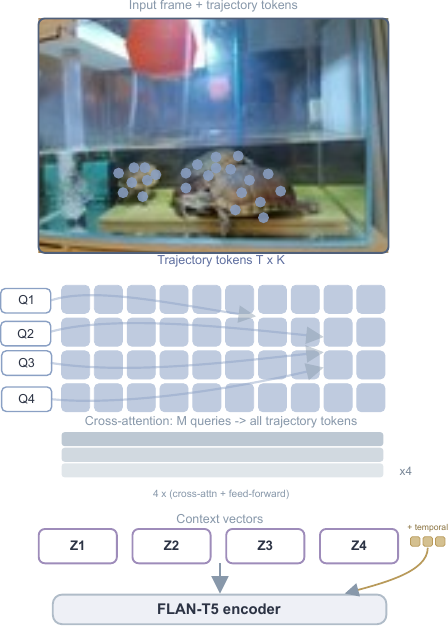}
  \caption{%
    Caption-Aware Resampler (\car).
    A video frame with its trajectory track tokens overlaid is shown at
    top; below, the flattened $T{\times}K$ token grid represents
    $\tilde{\Phi}$.
    $M$ learnable queries (Q1--Q4 shown) jointly cross-attend to the full
    grid over four stacked layers, via curved attention arcs whose
    thickness encodes attention weight; the queries are unordered
    compression slots, not object selectors, and \car\ itself produces no
    spatial grounding signal.
    Each query yields a context token (Z1--Z4); the $M$ tokens
    $\mathbf{Z}$ are projected and prepended, with per-frame temporal
    tokens, to the FLAN-T5 encoder.
    Spatial grounding is read out separately by the trajectory pointer
    head, which is not part of \car.
  }
  \label{fig:resampler}
\end{figure}%
}

\newcommand{\figQual}{%
\begin{figure*}[t]
  \centering
  \includegraphics[width=0.85\textwidth]{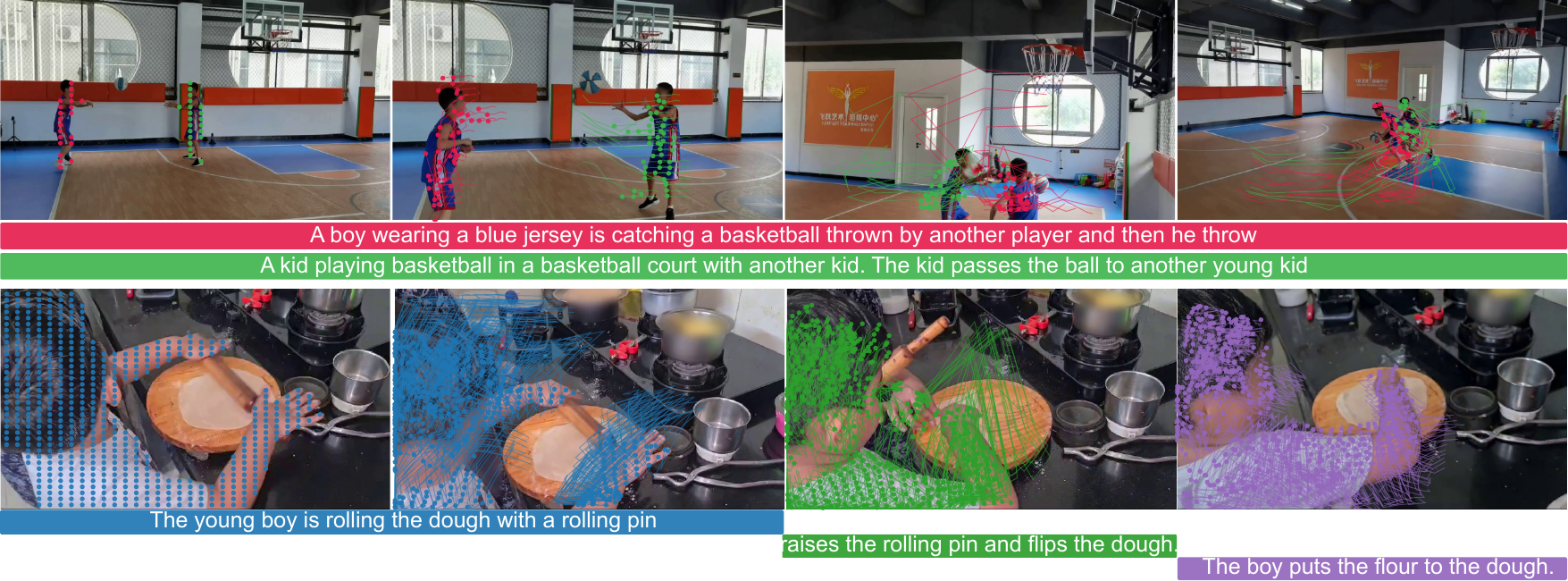}
  \caption{%
    Qualitative results on multi-subject and fine-grained motion.
    \textit{Top:} two players in a basketball rally, tracked and captioned
    independently and concurrently using visual and trajectory cues, with no query naming either.
    \textit{Bottom:} one subject performing fine-grained sub-actions, rolling
    dough, flipping it, adding flour, which \tcam\ segments into successive
    intervals with distinct pointers despite the overlapping hand motion.
  }
  \label{fig:qual}
\end{figure*}%
}

\newcommand{\figtrackingQual}{%
\begin{figure*}[t]
  \centering
  \includegraphics[width=\textwidth]{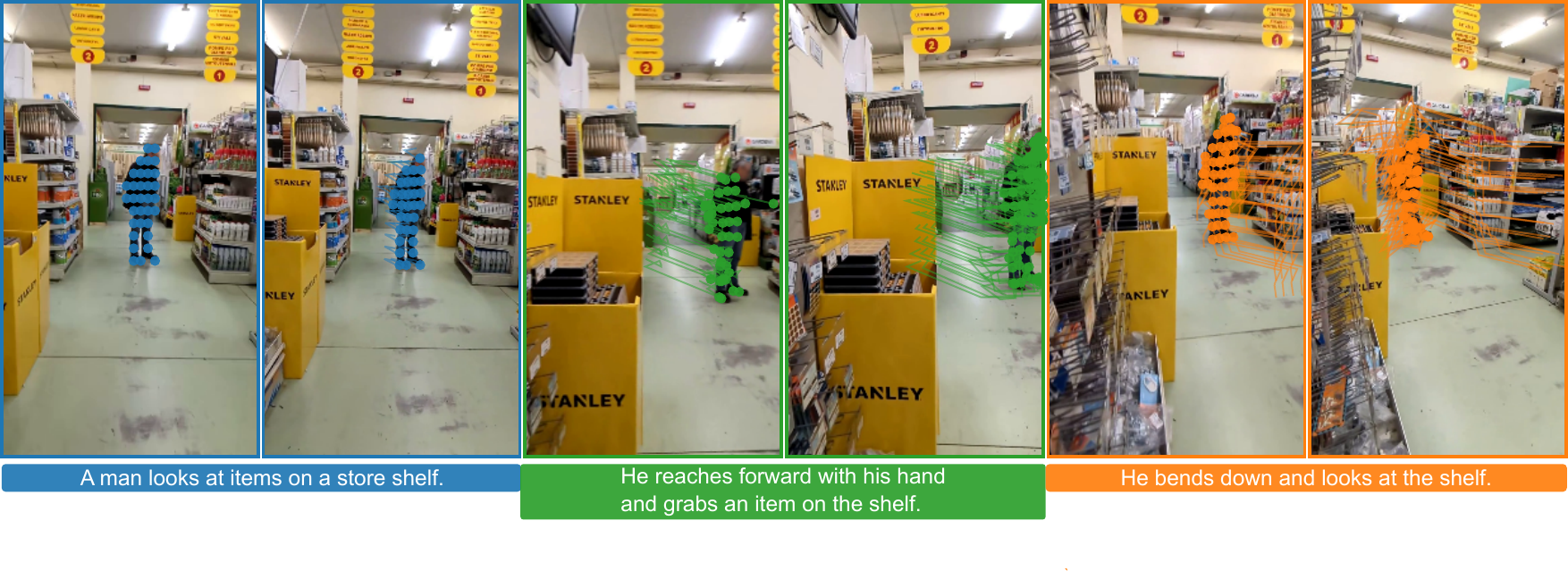}
  \caption{%
    Tracking robustness in a cluttered retail environment.
    A single subject is followed through a sequence of distinct
    sub-actions, looking at a shelf, reaching forward to grab an item, and
    bending down to inspect another shelf, while occluding store fixtures,
    densely packed products, and a moving viewpoint surround the subject
    throughout. \tcam\ maintains a stable trajectory pointer and an
    accurate caption through each pose change and partial occlusion,
    indicating that the trajectory tokens carry enough identity to survive
    the kind of visual clutter that often causes appearance-based trackers
    to drift or lose the target.
  }
  \label{fig:trackqual}
\end{figure*}%
}

\newcommand{\figSubObject}{%
\begin{figure*}[t]
  \centering
  \includegraphics[width=\textwidth]{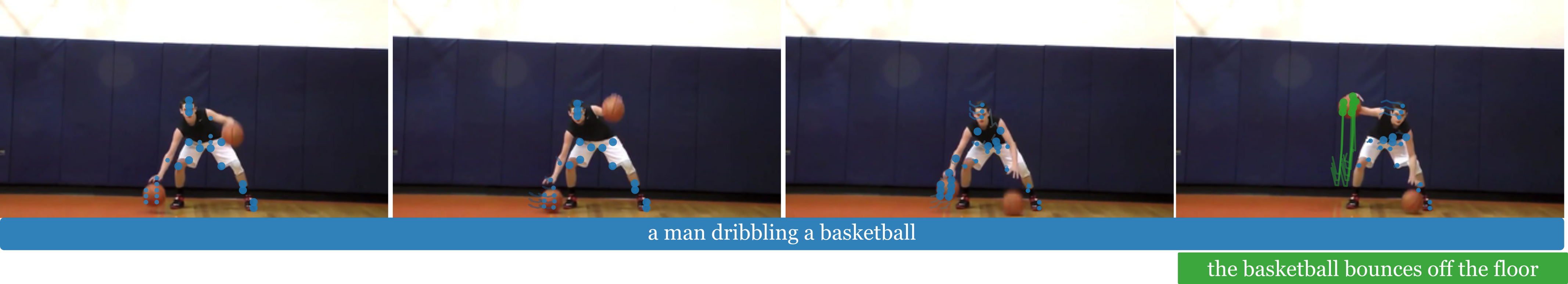}
  \vspace{-.05in}
  \caption{%
    Grounding below the object a detector would propose. A dribbling sequence,
    no query and no region prompt. \tcam\ holds the dribbler on one trajectory
    set (blue) and, at the bounce, splits the ball onto its own (green) with an
    independent caption and interval. This example illustrates separate attribution when the ball and person move differently, beyond assigning both motions to the person as a whole.
  }
  \label{fig:subobject}
\end{figure*}%
}

\definecolor{axAmber}{HTML}{B35E00}
\definecolor{axViolet}{HTML}{6A1B9A}
\definecolor{axBlue}{HTML}{2C5D8F}
\definecolor{axInk}{HTML}{2B3A82}

\newcommand{\tabGrounding}{%
\begin{table}[t]
\centering
\caption{Referring video grounding on Ref-YouTube-VOS~\cite{seo2020urvos} Val.
J\&F is the primary metric.
$^\dagger$ Uses the ground-truth text query at test time.}
\vspace{-.1in}
\label{tab:grounding}
\small
\setlength{\tabcolsep}{3pt}
\begin{tabular}{l|c|c|ccc}
\toprule
Method & Backbone & Query-free &
  \textcolor{axAmber}{J$\uparrow$} & \textcolor{axAmber}{F$\uparrow$} & \textcolor{axAmber}{J\&F$\uparrow$} \\
\midrule
ReferFormer$^\dagger$~\cite{wu2022referformer}  & Swin-L  & \xmark & 61.3 & 64.6 & 62.9 \\
UNINEXT$^\dagger$~\cite{yan2023uninext}          & ViT-H   & \xmark & \textbf{67.6} & \textbf{72.7} & \textbf{70.1} \\
OnlineRefer$^\dagger$~\cite{wu2023onlinerefer}   & Swin-L  & \xmark & 61.6 & 65.5 & 63.5 \\
\midrule
\textcolor{axInk}{\textbf{\tcam\ (Ours)}} & ViT-B/16 & \textcolor{axInk}{\cmark} &
  \textcolor{axInk}{59.8} & \textcolor{axInk}{61.2} & \textcolor{axInk}{60.5} \\
\bottomrule
\end{tabular}
\end{table}%
}

\newcommand{\tabCaptioning}{%
\begin{table}[t]
\centering
\caption{Dense video captioning on ActivityNet Captions~\cite{krishna2017dense}.
SODA\_c is the primary metric (jointly scores localization + caption quality).}
\vspace{-.1in}
\label{tab:captioning}
\small
\setlength{\tabcolsep}{4pt}
\begin{tabular}{l|ccc}
\toprule
Method & \textcolor{axViolet}{SODA\_c$\uparrow$} & \textcolor{axViolet}{CIDEr$\uparrow$} &
         \textcolor{axViolet}{METEOR$\uparrow$} \\
\midrule
PDVC~\cite{wang2021pdvc}                      & 6.0 & 29.3 & 7.6  \\
Vid2Seq~\cite{yang2023vid2seq}                & 5.9 & 30.2 & 8.5  \\
Streaming Vid2Seq~\cite{zhou2024streaming}    & 6.2 & 37.8 & 10.0 \\
\midrule
\textcolor{axInk}{\textbf{\tcam\ (Ours)}} &
  \textcolor{axInk}{6.5} & \textcolor{axInk}{39.4} & \textcolor{axInk}{10.6} \\
\bottomrule
\end{tabular}
\end{table}%
}

\newcommand{\tabTracking}{%
\begin{table}[t]
\centering
\caption{Trajectory pointer accuracy on TAP-Vid DAVIS and TAP-Vid
Kinetics~\cite{doersch2022tapvid}.
AJ: average Jaccard; OA: occlusion accuracy; $\delta_{avg}$: mean position accuracy over five thresholds. Dedicated trackers use all query points; TCAM uses a pointer-selected target subset with $k{=}32$. Its row is a selection diagnostic, not a like-for-like tracking comparison.}
\vspace{-.1in}
\label{tab:tracking}
\resizebox{\columnwidth}{!}{%
\begin{tabular}{l|ccc|ccc}
\toprule
\multirow{2}{*}{Method} &
  \multicolumn{3}{c|}{\textcolor{axBlue}{TAP-Vid DAVIS}} &
  \multicolumn{3}{c}{\textcolor{axBlue}{TAP-Vid Kinetics}} \\
& \textcolor{axBlue}{AJ$\uparrow$} & \textcolor{axBlue}{OA$\uparrow$} & \textcolor{axBlue}{$\delta_{avg}\uparrow$} &
  \textcolor{axBlue}{AJ$\uparrow$} & \textcolor{axBlue}{OA$\uparrow$} & \textcolor{axBlue}{$\delta_{avg}\uparrow$} \\
\midrule
PIPs++~\cite{zheng2023pointodyssey}    & --   & --   & 64.0 & --   & --   & --   \\
TAPIR~\cite{doersch2023tapir}          & 58.5 & 87.3 & 70.6 & 47.5 & 85.8 & 59.6 \\
CoTracker3~\cite{karaev2025cotracker3} & 64.5 & 90.9 & 77.1 & \textbf{54.4} & \textbf{89.4} & \textbf{66.0} \\
TAPNext~\cite{zholus2025tapnext}       & \textbf{66.6} & \textbf{92.2} & \textbf{79.5} & 53.0 & 89.3 & 64.5 \\
\midrule
\textcolor{axInk}{\textbf{\tcam\ pointer (Ours)}} &
  \textcolor{axInk}{64.8} & \textcolor{axInk}{90.4} & \textcolor{axInk}{76.8} &
  \textcolor{axInk}{53.9} & \textcolor{axInk}{88.9} & \textcolor{axInk}{65.7} \\
\bottomrule
\end{tabular}%
}
\end{table}%
}

\newcommand{\tabAblation}{%
\begin{table}[t]
\centering
\caption{Component ablation on MeViS Val (J\&F), ActivityNet Captions (METEOR),
and TAP-Vid DAVIS (AJ). Each row removes one module or training term from the
full model.}
\vspace{-.1in}
\label{tab:ablation}
\small
\setlength{\tabcolsep}{3pt}
\begin{tabular}{l|ccc}
\toprule
Configuration & \textcolor{axAmber}{J\&F$\uparrow$} &
                \textcolor{axViolet}{METEOR$\uparrow$} &
                \textcolor{axBlue}{AJ$\uparrow$} \\
\midrule
w/o Trajectory Tokenizer (ViT-only)   & 35.4 & 7.1  & --   \\
w/o VisionMotionEncoder               & 46.8 & 8.9  & 59.6 \\
w/o CAR (raw tokens) & 41.7 & 6.4  & 56.2 \\
w/o $\mathcal{L}_{grd}$               & 27.8 & 10.4 & 24.5 \\
w/o $\mathcal{L}_{div}$               & 42.5 & 10.5 & 62.0 \\
w/o structural token upweighting      & 44.9 & 8.2  & 63.1 \\
\midrule
\textcolor{axInk}{\textbf{\tcam\ (full)}} &
  \textcolor{axInk}{52.0} & \textcolor{axInk}{10.6} & \textcolor{axInk}{64.8} \\
\bottomrule
\end{tabular}
\end{table}%
}

\newcommand{\tabStructWeight}{%
\begin{table}[t]
\centering
\caption{Structural token upweight $w_{struct}$ ablation on MeViS Val.
Multi-event rate is the fraction of outputs containing $\geq 2$ events.}
\vspace{-.1in}
\label{tab:struct_weight}
\small
\begin{tabular}{c|cc}
\toprule
$w_{struct}$ & \textcolor{axAmber}{J\&F$\uparrow$} & \textcolor{axAmber}{Multi-event\%$\uparrow$} \\
\midrule
1.0          & 44.2 & 12.4 \\
\midrule
\textcolor{axInk}{\textbf{1.5 (ours)}} & \textcolor{axInk}{\textbf{52.0}} & \textcolor{axInk}{86.5} \\
\midrule
3.0          & 49.1 & 87.2 \\
5.0          & 46.3 & \textbf{88.0} \\
\bottomrule
\end{tabular}
\end{table}%
}

\newcommand{\tabDataScale}{%
\begin{table}[t]
\centering
\caption{Effect of training data scale on MeViS Val J\&F and
ActivityNet Captions METEOR.}
\vspace{-.1in}
\label{tab:data_scale}
\small
\begin{tabular}{l|cc}
\toprule
Training data & \textcolor{axAmber}{J\&F$\uparrow$} & \textcolor{axViolet}{METEOR$\uparrow$} \\
\midrule
MeViS only (1.6K)    & 44.8 & 7.2  \\
+ STGR (3.6K)        & 48.7 & 8.9  \\
\midrule
\textcolor{axInk}{\textbf{+ PLM (55K, ours)}} & \textcolor{axInk}{\textbf{52.0}} & \textcolor{axInk}{\textbf{10.6}} \\
\bottomrule
\end{tabular}
\end{table}%
}

\newcommand{\tabDataScaleRef}{\ref{tab:data_scale}}

\newcommand{\tabProtocol}{%
\begin{table}[t]
\centering
\caption{How a predicted event is matched to an annotated target. One set of
generations, re-scored under four selection rules. \tcam\ uses A; the oracle
gap is what a best-of-$N$ match would buy, and the query advantage is what
every baseline in Tab.~\ref{tab:grounding} is given for free. The last three
rows are degenerate pointers scored through the identical path, bounding what
the metric returns without a working pointer.}
\vspace{-.1in}
\label{tab:protocol}
\footnotesize
\setlength{\tabcolsep}{3.5pt}
\begin{tabular}{l|c|cc}
\toprule
\multirow{2}{*}{Event selection protocol} & Uses GT to &
  \multicolumn{2}{c}{\textcolor{axAmber}{J\&F$\uparrow$}} \\
& select? & MeViS & Ref-YT-VOS \\
\midrule
\textcolor{axInk}{\textbf{A. Confidence top-1 (ours)}} & \xmark\ query-free &
  \textcolor{axInk}{52.0} & \textcolor{axInk}{60.5} \\
B. Caption--expression match & text, post-hoc & 52.8 & 61.1 \\
C. Best-of-$N$ & \cmark\ \emph{oracle}   & 53.4 & 61.8 \\
D. Teacher-forced expression & \cmark\ query   & 56.2 & 64.5 \\
\midrule
\multicolumn{2}{l|}{\emph{oracle gap} (C\,--\,A)} & $+1.4$ & $+1.3$ \\
\multicolumn{2}{l|}{\emph{query advantage} (D\,--\,A)} & $+4.2$ & $+4.0$ \\
\multicolumn{2}{l|}{\emph{mean events / video}}   & 2.26 & 2.15 \\
\midrule
\multicolumn{2}{l|}{\emph{uniform pointer} (all 1296)}  & 3.0 & -- \\
\multicolumn{2}{l|}{\emph{random 40 tracks}}            & 3.7 & -- \\
\multicolumn{2}{l|}{\emph{fixed centre blob} ($r{=}18$)}  & 11.4 & -- \\
\bottomrule
\end{tabular}
\end{table}%
}

\newcommand{\tabCountPMR}{%
\begin{table}[t]
\centering
\caption{\textit{(a)} count correctness vs.\ the always-predict-one policy.
\textit{(b)} part-mass ratio; a mask-uniform pointer scores exactly $1.0$, so
$1.0$ is the uniform-attribution reference. $p$: paired Wilcoxon vs.\ the control.}
\vspace{-.1in}
\label{tab:countpmr}
\footnotesize
\setlength{\tabcolsep}{3pt}
\begin{tabular}{l|cccc}
\toprule
\multicolumn{5}{l}{\textbf{(a) Counting}\hfill P$\uparrow$\quad R$\uparrow$\quad F1$\uparrow$\quad MAE$\downarrow$} \\
\midrule
Event count (STGR) \ \textcolor{axInk}{\textbf{\tcam}} &
  \textcolor{axInk}{.947} & \textcolor{axInk}{.543} & \textcolor{axInk}{\textbf{.690}} & \textcolor{axInk}{1.92} \\
\quad always predict 1 & 1.000 & .254 & .405 & 2.94 \\
Subject count (MeViS) \ \textcolor{axInk}{\textbf{\tcam}} &
  \textcolor{axInk}{.882} & \textcolor{axInk}{.612} & \textcolor{axInk}{\textbf{.723}} & \textcolor{axInk}{1.24} \\
\quad always predict 1 & 1.000 & .318 & .483 & 2.14 \\
\midrule
\multicolumn{5}{l}{\textbf{(b) Sub-object attribution}\hfill split\quad med.\,PMR/null$\uparrow$\quad frac$>$null\quad $p$} \\
\midrule
\textcolor{axInk}{\textbf{mask BCE (\tcam)}} & contain &
  \textcolor{axInk}{1.32} & \textcolor{axInk}{.71} & -- \\
\quad $+$ softmax, uniform \emph{(control)} & contain & 1.32 & .71 & -- \\
\quad $+$ softmax, motion-weighted & contain & \textbf{1.44} & .78 & \textbf{.0018} \\
\midrule
\textcolor{axInk}{\textbf{mask BCE (\tcam)}} & moving &
  \textcolor{axInk}{\textbf{1.51}} & \textcolor{axInk}{.79} & -- \\
\quad $+$ softmax, uniform \emph{(control)} & moving & 1.51 & .78 & -- \\
\quad $+$ softmax, motion-weighted & moving & \textbf{1.69} & \textbf{.87} & \textbf{.0003} \\
\bottomrule
\end{tabular}
\end{table}%
}

\newcommand{\tabPointerFidelity}{%
\begin{table}[t]
\centering
\caption{Trajectory-pointer fidelity on all 30 TAP-Vid DAVIS videos, scored
against the \emph{same} frozen backbone the pointer reads from, so the readout
is isolated from the tracker. Chance is $k$ tracks drawn uniformly from the
same 1296. Chance is the backbone coverage, the fraction of the grid on target
and the chance rate in the limit $k\!\to\!K$, held at the measured constant
0.0398 for every row. The last row needs no $k$.}
\vspace{-.1in}
\label{tab:pointerfid}
\small
\begin{tabular}{l|ccc}
\toprule
Selection rule & \textcolor{axBlue}{acc.$\uparrow$} &
  \textcolor{axBlue}{chance} & \textcolor{axBlue}{ratio$\uparrow$} \\
\midrule
top-$k$, $k=8$    & .338 & .040 & \textbf{8.5}$\times$ \\
top-$k$, $k=16$   & .321 & .040 & 8.1$\times$ \\
top-$k$, $k=32$   & .306 & .040 & 7.7$\times$ \\
top-$k$, $k=64$   & .272 & .040 & 6.8$\times$ \\
top-$k$, $k=128$  & .211 & .040 & 5.3$\times$ \\
\midrule
\textcolor{axInk}{\textbf{threshold} $\sigma(s_{i,k})>0.005$} &
  \textcolor{axInk}{.294} & \textcolor{axInk}{.040} & \textcolor{axInk}{7.4$\times$} \\
\bottomrule
\end{tabular}
\end{table}%
}


\newcommand{\tabCamera}{%
\begin{table}[t]
\centering
\caption{Camera motion. Videos are split into tertiles by mean global
camera displacement; $\Delta$J\&F is measured against the low tertile.
Grounding \emph{improves} as the camera moves more.
\textit{Right:} adding an explicit residual-motion input (Eq.~\ref{eq:tau})
changes nothing, so no camera-compensation module is warranted.}
\vspace{-.1in}
\label{tab:camera}
\small
\setlength{\tabcolsep}{4pt}
\begin{tabular}{l|ccc|c}
\toprule
Camera tertile & $n$ & mean disp. & frac.\ moving &
  \textcolor{axAmber}{$\Delta$J\&F} \\
\midrule
low      & 17 & 0.003 & .25 & \emph{ref.} \\
medium   & 16 & 0.015 & .49 & $+2.6$ \\
high     & 17 & 0.040 & .57 & $\mathbf{+5.3}$ \\
\midrule
\multicolumn{4}{l|}{Pearson $r$(camera disp., J\&F)} & $+0.23$ \\
\multicolumn{4}{l|}{explicit compensation (matched, 1.5K steps)} & $+0.2$ \\
\bottomrule
\end{tabular}
\end{table}%
}

\newcommand{\tabDisruption}{%
\begin{table}[t]
\centering
\caption{Visual disruption. \emph{Disruption} is the mean frame-to-frame
appearance change; \emph{cuts} are detected shot boundaries. VISOR is
head-mounted egocentric footage, evaluated zero-shot. Grounding is unchanged
across a $3.8\times$ range of disruption, and half of the training videos
already contain a cut.}
\vspace{-.1in}
\label{tab:disruption}
\footnotesize
\setlength{\tabcolsep}{4pt}
\begin{tabular}{l|ccc|c}
\toprule
Dataset & $n$ & cuts/vid & disruption &
  \textcolor{axAmber}{$\Delta$J\&F} \\
\midrule
MeViS \texttt{valid\_u}          & 50 & 0.25 & 0.073 & \emph{ref.} \\
VISOR \emph{val} (egocentric)    & 43 & --   & \textbf{0.281} & $\mathbf{+0.3}$ \\
STGR/SA-V \emph{(train domain)}  & 40 & \textbf{0.62} & 0.141 & -- \\
\bottomrule
\end{tabular}
\end{table}%
}

\newcommand{\tabScalability}{%
\begin{table}[t]
\centering
\caption{\car\ scalability in track count. Same videos, same protocol, only the
CoTracker3 grid changes. A $4\times$ increase in $K$ costs $+60\%$ memory and no
extra latency, and accuracy is flat, because \car\ compresses any $K$ into the
same $M{=}64$ context tokens. V100-32GB, fp32, full per-video pass including
beam search.}
\vspace{-.1in}
\label{tab:scalability}
\small
\setlength{\tabcolsep}{4pt}
\begin{tabular}{l|c|ccc}
\toprule
Grid & $K$ & \textcolor{axAmber}{$\Delta$J\&F} &
  latency/vid & peak mem. \\
\midrule
$24^2$ & 576  & $-0.5$ & 1017\,ms & 2143\,MiB \\
\textcolor{axInk}{$\mathbf{36^2}$ \textbf{(ours)}} & \textcolor{axInk}{1296} &
  \textcolor{axInk}{\emph{ref.}} & \textcolor{axInk}{909\,ms} & \textcolor{axInk}{2592\,MiB} \\
$48^2$ & 2304 & $+0.4$ & 871\,ms  & 3435\,MiB \\
\bottomrule
\end{tabular}
\end{table}%
}

\newcommand{\tabTrackerSwap}{%
\begin{table}[t]
\centering
\caption{Backbone dependence. The frozen trajectory backbone is replaced by an
unrelated tracker family (cost-volume matching instead of joint transformer
attention). \textit{Top:} no retraining, pointer-selection accuracy on TAP-Vid
DAVIS. \textit{Bottom:} two matched 2K-step retrainings.}
\vspace{-.1in}
\label{tab:trackerswap}
\small
\setlength{\tabcolsep}{4pt}
\begin{tabular}{l|ccccc}
\toprule
\multicolumn{6}{l}{\textbf{(a) No retraining} \hfill accuracy at $k=$} \\
& 8 & 16 & 32 & 64 & 128 \\
\midrule
CoTracker3~\cite{karaev2025cotracker3} & .338 & .321 & .306 & .272 & .211 \\
BootsTAPIR~\cite{doersch2024bootstap}  & .341 & .315 & .286 & .261 & .216 \\
$\Delta$ & \multicolumn{5}{c}{$|\Delta| \leq 0.021$ at every $k$; sign flips} \\
\midrule
\multicolumn{6}{l}{\textbf{(b) Matched retraining} \hfill $\Delta$J\&F \quad event-F1 \quad MAE} \\
\midrule
\multicolumn{3}{l}{CoTracker3} & \emph{ref.} & .690 & 1.92 \\
\multicolumn{3}{l}{BootsTAPIR} & $+0.2$ & \textbf{.695} & \textbf{1.90} \\
\bottomrule
\end{tabular}
\end{table}%
}

\newcommand{\tabStructPlateau}{%
\begin{table}[t]
\centering
\caption{$w_{struct}$ under a \emph{matched} protocol: six 600-step finetunes
from the released checkpoint, identical in every respect except
$w_{struct}$. Both response surfaces are flat, including at the three points
Tab.~\ref{tab:struct_weight} samples, so the operating point is a plateau
rather than a tuned peak. The multi-event collapse there is a property of
training from scratch at $w_{struct}{=}1.0$, which a short finetune cannot
undo.}
\vspace{-.1in}
\label{tab:structplateau}
\footnotesize
\setlength{\tabcolsep}{4pt}
\begin{tabular}{c|ccc}
\toprule
$w_{struct}$ & \textcolor{axAmber}{$\Delta$J\&F} &
  event-F1$\uparrow$ & mean pred.\ events \\
\midrule
1.00 & $-0.4$ & .675 & 2.12 \\
1.25 & $-0.2$ & .684 & 2.20 \\
\textcolor{axInk}{\textbf{1.50 (ours)}} & \textcolor{axInk}{\emph{ref.}} &
  \textcolor{axInk}{.690} & \textcolor{axInk}{2.26} \\
1.75 & $+0.1$ & .686 & 2.24 \\
2.00 & $+0.1$ & .681 & 2.22 \\
3.00 & $-0.3$ & .678 & 2.20 \\
\midrule
\emph{total range} & \textbf{0.5} & \textbf{.015} & 0.14 \\
\bottomrule
\end{tabular}
\end{table}%
}

\newcommand{\tabPMRmatched}{%
\begin{table}[t]
\centering
\caption{Sub-object selectivity at a real budget: five \emph{matched} 3K-step
finetunes on the moving-part split ($n{=}105$ pairs), each motion-weighted arm
paired with a uniform control at the \emph{same} $\lambda_{sm}$, so the effect
is attributable to the weighting and not to the extra term. Raising
$\lambda_{sm}$ with a uniform target is flatly null ($p{=}.782$); only the
motion-weighted target moves the metric. Bonferroni threshold over the eight
tests run is $.00625$, which all three motion-weighted arms clear.}
\vspace{-.1in}
\label{tab:pmrmatched}
\footnotesize
\setlength{\tabcolsep}{4pt}
\begin{tabular}{l|c|cc|c}
\toprule
Target & $\lambda_{sm}$ & med.\,PMR/null$\uparrow$ & frac$>$null & $p$ \\
\midrule
uniform \emph{(control)}   & 1.0 & 1.215 & .657 & -- \\
\textcolor{axInk}{motion-weighted} & 1.0 & \textcolor{axInk}{1.528} &
  \textcolor{axInk}{.790} & \textcolor{axInk}{\textbf{.00112}} \\
uniform \emph{(control)}   & 4.0 & 1.220 & .667 & .782 \\
\textcolor{axInk}{motion-weighted} & 4.0 & \textcolor{axInk}{\textbf{1.685}} &
  \textcolor{axInk}{\textbf{.867}} & \textcolor{axInk}{\textbf{.00014}} \\
\textcolor{axInk}{motion-weighted, $p{=}2$} & 4.0 & \textcolor{axInk}{\textbf{1.742}} &
  \textcolor{axInk}{\textbf{.895}} & \textcolor{axInk}{\textbf{.00008}} \\
\bottomrule
\end{tabular}
\end{table}%
}

\newcommand{\tabThreshold}{%
\begin{table}[t]
\centering
\caption{Replacing top-$k$ with a threshold on the pointer posterior. The
selected set size is chosen by the model, not by the evaluator, and
$\tau = 0.005$ matches fixed $k{=}32$ while referring to no $k$ at all.}
\vspace{-.1in}
\label{tab:threshold}
\small
\begin{tabular}{c|cc}
\toprule
$\tau$ & mean $|$selected$|$ & \textcolor{axBlue}{accuracy$\uparrow$} \\
\midrule
0.001 & 202.3 & .223 \\
0.002 & 138.6 & .254 \\
\textcolor{axInk}{\textbf{0.005}} & \textcolor{axInk}{55.2} & \textcolor{axInk}{\textbf{.294}} \\
\bottomrule
\end{tabular}
\end{table}%
}

\twocolumn[{
\renewcommand\twocolumn[1][]{#1}
\maketitle
\begin{center}
    \vspace{-28pt}
    \includegraphics[width=0.95\linewidth]{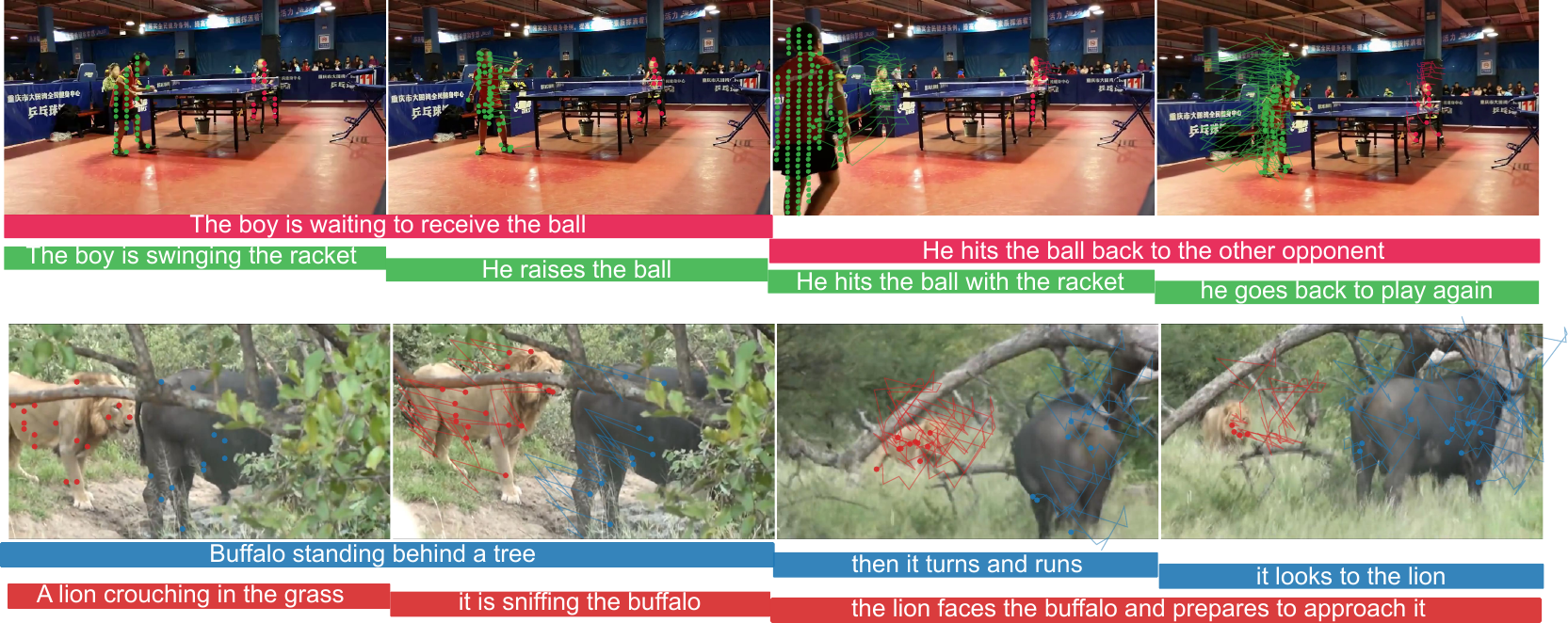}
    \centering
    \captionsetup{type=figure}
    \vspace{-5pt}
    \caption{%
      \tcam\ watches raw video with no text query and no region prompt, and
      produces temporally ordered, open-vocabulary captions grounded in dense
      point trajectories. \textit{Top:} a table-tennis rally with two
      concurrent subjects, each on its own trajectory set (red, green) with a
      separate caption stream. \textit{Bottom:} a lion approaching a buffalo,
      associated with distinct trajectory sets. Neither subject is named in advance: the
      model describes their motion using trajectory and visual features.
    }
    \label{fig:teaser}
\end{center}
}]

\begin{abstract}
We present \tcam\ (Track and Caption Any Motion), a generative framework that watches video without a text query or region prompt and produces open-vocabulary captions, temporal intervals, and pointers to the trajectories supporting each event. Dense point trackers localize motion without describing it, while clip-level video-language representations do not directly identify the trajectories carrying an event. \tcam\ connects these capabilities through a Caption-Aware Resampler: learned queries compress dense trajectory tokens into a fixed-length context for a language decoder. One autoregressive pass describes sequential events and concurrent subjects, each with its own trajectory pointer. Training uses existing captions and segmentation annotations, without additional event labeling. With over 50K training clips, \tcam\ outperforms the evaluated dense-captioning baselines on ActivityNet Captions and approaches query-based grounding methods on Ref-YouTube-VOS, without training on either dataset. Event selection uses model confidence; best-of-$N$ selection improves grounding by at most $1.4$ J\&F points in the reported comparisons. Pointer selection exceeds chance by $5.3$--$8.5\times$, and pointer mass concentrates on moving sub-parts above a mask-uniform reference. These selection diagnostics are distinct from full-query tracking accuracy.
Code is available at \codelink.
\end{abstract}
\vspace{-20pt}
\section{Introduction}
\label{sec:intro}

\figLandscape

Imagine watching a wildlife video: a lion moves through tall grass while a buffalo
stands alert in the background (Fig.~\ref{fig:teaser}, bottom). Without any instruction, a human instantly
registers that two distinct motion events are unfolding, the stalking and the
vigilant stillness, describes each one, locates it in time, and points to the
precise trajectories that carry its meaning. Crucially, what separates the two
events is \emph{motion}, not appearance: freeze any single frame and the story
collapses. This effortless fusion of tracking, temporal segmentation, and
language remains out of reach for current video understanding systems, and the
reason is structural: the field is split along the very axes a complete answer
must span (Fig.~\ref{fig:landscape}).

Dense point trackers~\cite{karaev2025cotracker3, doersch2023tapir, harley2022pips}
follow pixels with remarkable precision, but a track is a curve, not a sentence:
they localize without ever interpreting. Video-language
models~\cite{luo2022clip4clip, ma2022xclip, wu2022referformer, yan2023uninext} go
the other way, connecting vision to text, but only once the user supplies a
query, and so they answer questions rather than discover events. Dense captioning
methods~\cite{yang2023vid2seq, wang2021pdvc} automate event description, yet they
reason from clip-level features and are blind to which pixels actually move.
A handful of object-level captioners~\cite{zhou2023dense, fiastre2025maskcaptioner}
do pair tracking with language, but they ride on detector boxes or segmentation
masks and inherit \emph{object} granularity: they never reach the individual
trajectories that define fine-grained motion, and they describe objects a detector
already proposes rather than motion a model discovers. Each line of work owns one
or two axes of Fig.~\ref{fig:landscape}; none owns all three.

The gap is therefore precise. No existing model jointly (1) localizes at
\emph{point-trajectory} granularity, (2) generates temporally ordered,
open-vocabulary multi-event captions, and (3) grounds each caption in the
trajectories that produced it, all with no text query and no region prompt.
\tcam\ occupies exactly this empty corner.

\tcam\ takes raw video and produces a sequence of structured events, each a
free-form caption, a temporal interval, and a trajectory pointer. A frozen
CoTracker3 backbone supplies dense point trajectories. A lightweight
\textit{Caption-Aware Resampler} (CAR) distills these trajectory tokens into a
fixed-length motion context through learned cross-attention queries. A FLAN-T5~\cite{chung2022flan}
decoder conditioned on this context emits the full event sequence in one
autoregressive pass, using special tokens to delimit boundaries and fire pointer
activations. For the lion-and-buffalo clip, this produces two interleaved
streams, one per subject, with overlapping intervals:

\vspace{1pt}
\texttt{\small\textbf{<CAP\_BEGIN>} <T\textsubscript{0}><T\textsubscript{9}>
\textit{lion crouching in grass} \textbf{<TRK>} \textbf{<CAP\_END>}
\textbf{<CAP\_BEGIN>} <T\textsubscript{7}><T\textsubscript{15}>
\textit{buffalo turns and runs} \textbf{<TRK>} \textbf{<CAP\_END>} \ldots}
\vspace{1pt}

Each \texttt{<TRK>} fires a lightweight pointer that scores the same dense
trajectory tokens the caption is generated from, so grounding is not a separate
detection stack: it rides on the one backbone that decides what to describe. Training uses a three-term objective that supervises
caption quality, spatial pointer alignment against object masks, and trajectory
diversity, derived entirely from existing segmentation annotations and requiring
no manual event labeling. Scaling to PLM-Video-Human~\cite{cho2025perceptionlm}
(55K videos, 117K samples) confirms that the architecture absorbs more data
without task-specific engineering.

Our contributions are:
\begin{itemize}\itemsep1pt
\item \textbf{Event discovery as generation, not retrieval.} Motion
  understanding is recast as one autoregressive pass that decides \emph{how
  many} events a video holds and emits each, instead of retrieving a span
  against a query or set-predicting over detector proposals. The count is an
  output, and we measure it: event- and subject-count F1 far exceed the
  degenerate one-event policy (\S\ref{sec:counting}).

\item \textbf{Language conditioned on trajectories, not clips.} Conditioning on
  dense tracks compressed to a fixed-length context lets a caption follow from
  \emph{how} pixels move. The compression is what makes it practical: the
  language model never sees $K$, so track density is free to grow
  (\S\ref{sec:robust}).

\item \textbf{Grounding as a readout of the generator's own evidence.} The
  pointer scores the same tokens that produced the caption, so spatial support
  explains the model's decision rather than coming from a second stack. It
  selects on-target tracks $5.3$--$8.5\times$ above chance from a grid where only
  $4\%$ are on target (\S\ref{sec:tapvid}).

\item \textbf{Mask supervision is a floor, not a ceiling.} All of this trains
  from existing masks with no event or point labels, yet the granularity is not
  the mask's: pointer mass concentrates on moving sub-parts at
  $1.5$--$1.7\times$ the mask-uniform area-ratio reference (\S\ref{sec:counting}).
\end{itemize}

Evaluated in-domain on MeViS and by transfer on ActivityNet Captions and TAP-Vid, \tcam\ is, to our knowledge, the first model to deliver point-level
grounding, temporal structure, and open-vocabulary generation in a single
query-free pass: tracking and language, together, at motion granularity.

\section{Related Work}
\label{sec:related}

\tcam\ sits at the intersection of point tracking, video-language models, dense
video captioning, and generative multimodal reasoning. Each field has advanced
rapidly, yet none has produced a model that unifies spatial trajectory precision
with open-vocabulary, query-free generation.

\noindent\textbf{Dense Point Tracking.}
Modern point trackers follow pixels with exceptional accuracy over long
sequences. CoTracker~\cite{karaev2024cotracker} and
CoTracker3~\cite{karaev2025cotracker3} track thousands of points jointly with
transformer architectures and virtual tracks for occlusion handling;
PIPs~\cite{harley2022pips} and PIPs++~\cite{zheng2023pointodyssey} treat pixels
as independent particles refined over time; TAPIR~\cite{doersch2023tapir} and
BootsTAPIR~\cite{doersch2024bootstap} use two-stage matching and refinement,
while TAPNext~\cite{zholus2025tapnext} recasts tracking as next-token
prediction. 3D extensions~\cite{xiao2024spatialtracker,zhang2025tapip3d} lift trajectories
to geometry, and TAP-Vid~\cite{doersch2022tapvid} standardizes evaluation. All
are purely geometric: they track points but cannot interpret, segment, or
describe the motion they observe. \tcam\ uses CoTracker3 as a frozen backbone
for its trajectory quality but conditions language generation directly on the
resulting tokens, and \S\ref{sec:robust} shows the coupling is not specific to
that tracker.
\figPipeline

\noindent\textbf{Video-Language Models.}
CLIP-based methods~\cite{radford2021clip, luo2022clip4clip, ma2022xclip,
xu2021videoclip, wang2021actionclip} learn joint video-text embeddings;
spatio-temporal grounding~\cite{yang2022tubedetr, jin2022stcat, gu2024cgstvg}
localizes text-described objects with transformer decoders; referring video
segmentation~\cite{wu2022referformer, yan2023uninext, wang2024maris} achieves
pixel-level grounding given an expression. One limitation runs through all of
them: without a user-supplied query there is no retrieval, no grounding, no
output. \tcam\ requires none, discovering and grounding events from motion
alone.

\noindent\textbf{Dense and Object-Level Video Captioning.}
Dense captioning partitions a video into temporal segments and describes each.
PDVC~\cite{wang2021pdvc} casts this as parallel set prediction with a DETR-like
head; Vid2Seq~\cite{yang2023vid2seq} pretrains on 18M narrated videos with time
tokens for boundary prediction. Both locate events in time but reason from
clip-level features, so a caption is never tied to the pixels that produced it.
The closest work adds spatial identity: E2ESG~\cite{zhou2023dense} captions
object trajectories from detection-based tracking,
MaskCaptioner~\cite{fiastre2025maskcaptioner} jointly segments and captions
them, and streaming variants~\cite{zhou2024streaming} target online settings.
These are the right idea at the wrong granularity: they describe \emph{objects}
a detector proposes, with no mechanism to attribute a caption to specific
trajectories. \tcam\ grounds every caption in a pointer over the dense grid,
and \S\ref{sec:counting} shows this reaches sub-object motion a mask cannot
represent.

\noindent\textbf{Generative Models for Video Understanding.}
VideoChat~\cite{li2023videochat} and VideoLLaMA align video encoders to LLMs
for open-ended dialogue, TimeChat~\cite{ren2024timechat} adds timestamp-aware
tokens, and VTimeLLM~\cite{huang2024vtimellm} localizes events through
instruction tuning. All rely on clip-level features: they do not track points,
cannot produce trajectory pointers, and cannot resolve which trajectories carry
a particular event.

\noindent\textbf{Positioning TCAM.}
Each prior family owns a face of the design space; \tcam\ is the first to reach
the corner where all three axes meet (Fig.~\ref{fig:landscape}). It keeps the
point-trajectory granularity of trackers~\cite{karaev2025cotracker3} but adds
language; it produces the spatial grounding of referring
segmentation~\cite{wu2022referformer} while discovering events without a
query~\cite{yang2022tubedetr}; it generates the open-vocabulary descriptions of
dense captioners~\cite{ren2024timechat, huang2024vtimellm} at point rather than
clip granularity; and where object-level captioners~\cite{zhou2023dense,
fiastre2025maskcaptioner} describe detector-proposed objects, \tcam\ describes
and grounds motion directly from dense trajectories.
\section{TCAM: Track and Caption Any Motion}
\label{sec:method}

\subsection{Problem Formulation}
\label{sec:formulation}

Given a video $\mathbf{V} = \{\mathbf{I}_t\}_{t=1}^{T}$ with $T$ frames and
resolution $H \times W$, we seek to produce a set of structured events
$\mathcal{E} = \{e_i\}_{i=1}^{N}$, where each event is a triple
\begin{equation}
  e_i = \bigl(\,\text{caption}_i,\;\; [f_s^i, f_e^i],\;\; \hat{\mathbf{p}}_i\,\bigr),
  \label{eq:event}
\end{equation}
consisting of a natural language description, a temporal interval
$[f_s^i, f_e^i] \subset [1, T]$, and a trajectory pointer
$\hat{\mathbf{p}}_i \in \mathbb{R}^{K}$ that assigns relevance weights over $K$
tracked points. The number of events $N$ is not fixed: the model determines it
from motion content alone. The set $\mathcal{E}$ is jointly \emph{multi-temporal}
and \emph{multi-spatial}: events may be ordered sequentially in time
(distinct intervals of the same or different subjects) and may also co-occur,
with several subjects active in the same interval, each carrying its own
trajectory pointer $\hat{\mathbf{p}}_i$. No user text query is provided at any
stage.

\noindent\textbf{What counts as moving.} Because no query names the subject,
``moving'' must be defined against the camera rather than the image. For each
frame pair we fit a global 2D affine transform $\mathbf{H}_t$ by RANSAC over
all $K$ tracks, the background being the dominant inlier set, and take a
track's \emph{residual} displacement
$\mathbf{x}^{res}_{j,t} = \mathbf{x}_{j,t} - \mathbf{H}_t(\mathbf{x}_{j,t-1})$.
A track is moving when its residual speed exceeds
$\text{med}(\text{inliers}) + 2\,\text{MAD}(\text{inliers})$. Object motion is
therefore separated from camera motion by construction, and the quantity is
measurable, which lets us stratify results by camera motion in
\S\ref{sec:robust}.

\subsection{Architecture}
\label{sec:arch}

\tcam\ is built from three sequential modules: a \textit{Trajectory
Tokenizer}, a \textit{Caption-Aware Resampler} (CAR), and a \textit{Structured
Generator}. Figure~\ref{fig:pipeline} illustrates the full pipeline.

\noindent\textbf{Trajectory Tokenization.}
We extract dense point trajectories using frozen CoTracker3~\cite{karaev2025cotracker3}
initialized on a $36 \times 36$ regular grid ($K = 1296$ tracks). For each
track $j$ at frame $t$, CoTracker3 provides position $\mathbf{x}_{j,t} \in
\mathbb{R}^2$, visibility $v_{j,t} \in \{0,1\}$, and the velocity
$\dot{\mathbf{x}}_{j,t} = \mathbf{x}_{j,t} - \mathbf{x}_{j,t-1}$.
Three lightweight MLPs encode these signals independently:
\begin{equation}
  \phi_{j,t} = \text{MLP}_{pos}(\mathbf{x}_{j,t}) +
               \text{MLP}_{vel}(\dot{\mathbf{x}}_{j,t}) +
               \text{MLP}_{vis}(v_{j,t}).
  \label{eq:track_tokens}
\end{equation}
The $T \times K$ track tokens $\Phi \in \mathbb{R}^{T \times K \times d}$ are
then enriched by cross-attention against frozen ViT-B/16 frame features
$\mathbf{F} = \{\mathbf{f}_t\}_{t=1}^{T}$, producing appearance-aware motion
tokens $\tilde{\Phi}$ (the \textit{VisionMotionEncoder} block). Both backbones stay frozen: a trainable one could minimize the caption loss by
drifting toward appearance shortcuts, recognizing what an object looks like
rather than how it moves, eroding the motion signal the trajectory tokens exist
to supply. Freezing forces task-specific learning into the resampler and
decoder, so appearance enriches the trajectory representation without
overwriting it. Nothing here is specific to CoTracker3, and \S\ref{sec:robust}
confirms the readout transfers to an unrelated tracker family.

\noindent\textbf{Caption-Aware Resampler (CAR).}
\label{sec:car}

The $T \times K$ trajectory tokens are too large to feed directly into a
sequence model. CAR compresses them to a fixed-length context through $M$
learnable query vectors $\mathbf{Q}^{(0)} \in \mathbb{R}^{M \times d}$ that
are refined over $L = 4$ stacked layers, each a cross-attention against
$\tilde{\Phi}$ followed by a feed-forward sub-layer:
\begin{equation}
  \mathbf{Q}^{(\ell)} = \text{FFN}\Bigl(\text{CrossAttn}\bigl(\mathbf{Q}^{(\ell-1)},\; \tilde{\Phi}\bigr)\Bigr),
  \qquad \ell = 1, \ldots, L,
  \label{eq:car}
\end{equation}
with the final layer's output taken as the motion context vectors,
$\mathbf{Z} = \mathbf{Q}^{(L)} \in \mathbb{R}^{M \times d}$. The design is
deliberately minimal and leaves the FLAN-T5 weights unmodified apart from LoRA
adapters. Because the cross-attention operates over all $T \times K$
trajectory tokens jointly, each context vector integrates information across
time and space, making it sensitive to spatiotemporal motion rather than
single-frame appearance. Fixing the output at $M$ vectors also decouples the language model
from $K$, which is what makes track density cheap (\S\ref{sec:robust});
supp.\ Fig.~\ref{fig:resampler} illustrates the resampler in detail.
The $M$ context tokens $\mathbf{Z}$ are projected to the FLAN-T5 hidden size and
prepended to its \emph{encoder} input, alongside $T$ \emph{per-frame temporal
tokens}, one spatially-pooled token per sampled frame with a learned
frame-position embedding. These restore the frame identity the unordered
$\mathbf{Z}$ tokens discard, which is what lets the model place events in time
via the \texttt{<T$_i$>} tokens below. The decoder cross-attends to these
encoder states to generate captions; spatial grounding is read out separately
by a lightweight \emph{trajectory pointer} sharing the same representation, so
both tasks run on one backbone rather than separate stacks.

\noindent\textbf{Structured Generation.}
We adapt FLAN-T5-base~\cite{chung2022flan} as the language decoder.
We add special tokens to the vocabulary: \texttt{<CAP\_BEGIN>} (event
start), \texttt{<TRK>} (trajectory pointer trigger), \texttt{<CAP\_END>}
(event end), and a set of discrete \emph{time tokens} \texttt{<T$_0$>}\,\ldots\,\texttt{<T$_{T-1}$>},
one per sampled frame, that mark each event's temporal interval in the
Vid2Seq~\cite{yang2023vid2seq} style. Their rows in the output embedding matrix (\texttt{lm\_head}) are unfrozen while
all other vocabulary rows stay frozen, via a gradient hook that zeroes
gradients at non-special positions. LoRA adapters~\cite{hu2022lora} with
$r = 16$ are applied to the Q, K, V projections of all T5 attention layers.

At inference, the decoder generates the full structured sequence
autoregressively, cross-attending to the encoder states (the context tokens
$\mathbf{Z}$ and temporal tokens):
\begin{align}
  &\texttt{<CAP\_BEGIN>}\; \texttt{<T}_{f_s^1}\texttt{>}\texttt{<T}_{f_e^1}\texttt{>}\; \text{caption}_1\; \texttt{<TRK>}\; \texttt{<CAP\_END>}\;\nonumber\\
  &\texttt{<CAP\_BEGIN>}\; \texttt{<T}_{f_s^2}\texttt{>}\texttt{<T}_{f_e^2}\texttt{>}\; \text{caption}_2\; \texttt{<TRK>}\; \texttt{<CAP\_END>}\;\ldots
  \label{eq:output}
\end{align}
Each \texttt{<TRK>} token triggers the trajectory pointer $\hat{\mathbf{p}}_i$:
its decoder hidden state queries the $K$ enriched track tokens $\tilde{\Phi}$
through a two-layer cross-attention head with a bilinear readout, yielding a
distribution over the $K$ tracks. The pointer is a lightweight head rather than
a separate detector, consuming the same vision-motion tokens that condition the
caption. Because every clause carries its own \texttt{<TRK>}, one pass expresses both
temporally ordered events and concurrent subjects: successive clauses may
describe different intervals of one subject or different subjects active in the
same window, each with an independent pointer. The number of events is
unconstrained, set by when the decoder emits its final \texttt{<CAP\_END>}.

\subsection{Training Objectives}
\label{sec:loss}

We train with three complementary terms:
\begin{equation}
  \mathcal{L} = \mathcal{L}_{cap} + \lambda_{grd}\,\mathcal{L}_{grd}
                                   + \lambda_{div}\,\mathcal{L}_{div}.
  \label{eq:total_loss}
\end{equation}

\noindent\textbf{Caption loss $\mathcal{L}_{cap}$.}
Label-smoothed cross-entropy over the target token sequence. Structural tokens
(\texttt{<CAP\_BEGIN>}, \texttt{<TRK>}, \texttt{<CAP\_END>}) are upweighted
by a factor $w_{struct}$ to ensure the model learns to produce them reliably;
ablations (\S\ref{sec:ablations}) show this is critical for multi-event generation.

\noindent\textbf{Grounding loss $\mathcal{L}_{grd}$.}
The trajectory pointer should score tracks that fall inside the target object
highly. Given the per-frame binary mask $\mathbf{m}_t \in \{0,1\}^{H \times W}$
for event $i$, we take the union mask over its interval $[f_s^i, f_e^i]$ and
mark every track whose position lands inside it, giving a binary on-object
target $y_{i,k} \in \{0,1\}$ over the $K$ tracks. The pointer logits
$s_{i,k}$ are then supervised with a class-balanced binary cross-entropy:
\begin{equation}
  \mathcal{L}_{grd} = \sum_i \sum_{k=1}^{K}
    \text{BCE}\bigl(\,\sigma(s_{i,k}),\; y_{i,k}\,\bigr).
  \label{eq:l_grd}
\end{equation}
This supervision is derived entirely from existing segmentation annotations;
no additional labeling is required.

A binary on-object target is uniform inside the mask, so on its own it can only
teach the pointer to select the object. To let the same masks express
\emph{which part of} the object moves, we add a softmax competition term whose
target is graded rather than flat: each on-object track is weighted by its
residual displacement, $q_{i,k} \propto y_{i,k}\,\lVert \mathbf{x}^{res}_{k}
\rVert^{\,p}$, after the object's own median rigid displacement is removed. The
positive set is unchanged, so ``point somewhere on the object'' is preserved;
only the competition among on-object tracks changes. This costs no new
annotation and measurably sharpens sub-object attribution
(\S\ref{sec:counting}).

\noindent\textbf{Diversity loss $\mathcal{L}_{div}$.}
When a video contains multiple events, whether sequential intervals or
concurrent subjects, the pointer vectors should be distinct.
We penalize redundancy with a pairwise cosine similarity loss:
\begin{equation}
  \mathcal{L}_{div} = \sum_{i \neq j} \max\!\bigl(0,\;
    \cos(\hat{\mathbf{p}}_i, \hat{\mathbf{p}}_j) - \delta\bigr),
  \label{eq:l_div}
\end{equation}
where $\delta$ is a margin that tolerates partial overlap between events.
Without this term, the model tends to produce identical pointers for distinct
events that share the same spatial region.
\section{Experiments}
\label{sec:experiments}

\subsection{Experimental Setup}
\label{sec:setup}

\noindent\textbf{Datasets.}
We train on three sources: the MeViS training split~\cite{ding2023mevis}
(1,662 videos, 27,502 motion expressions with masks), the STGR subset of
PLM-Video-Human with dense multi-event captions (2,048 videos), and the full
PLM-Video-Human rdcap split~\cite{cho2025perceptionlm} (55,002 videos, 117,248
samples). Training warms up on MeViS + STGR, then continues on the full
collection; \S\ref{sec:ablations} isolates each contribution. No other corpus
enters training: neither ActivityNet Captions nor Ref-YouTube-VOS is trained
on, so \tcam's rows in Tables~\ref{tab:grounding} and~\ref{tab:captioning} are
\emph{zero-shot transfer} while every baseline they face is trained
in-domain.

\noindent\textbf{Implementation.}
Trajectories come from frozen CoTracker3~\cite{karaev2025cotracker3} at
$T = 16$ frames and $K = 1296$ tracks ($36 \times 36$ grid), with frozen
ViT-B/16~\cite{radford2021clip} as the visual backbone, $M = 64$ resampler
queries, and a FLAN-T5-base~\cite{chung2022flan} decoder adapted with LoRA
($r = 16$ on Q, K, V). We set $\lambda_{grd} = 1.0$, $\lambda_{div} = 0.1$,
$w_{struct} = 1.5$ and $\delta = 0.3$, and train with AdamW at
$5\!\times\!10^{-5}$, batch size 16 with gradient accumulation over 2 steps,
for 18K steps on one H100 with bf16.

\noindent\textbf{Evaluation.}
We evaluate the three capabilities on the benchmark native to each:
\textit{spatiotemporal grounding} on MeViS Val and Ref-YouTube-VOS Val
(J, F, J\&F); \textit{dense video captioning} on ActivityNet
Captions~\cite{krishna2017dense} (SODA\_c, CIDEr, METEOR); and
\textit{trajectory pointer accuracy} on TAP-Vid DAVIS and
Kinetics~\cite{doersch2022tapvid} (AJ, OA, $\delta_{avg}$).

\noindent\textbf{Which predicted event is scored.}
\label{sec:protocol}
Every benchmark above scores one target per annotation, while \tcam\ emits $N$
events with no query and no canonical ordering, so the rule pairing the two
decides what the numbers mean. We use \textbf{confidence top-1}: events are
ranked by the mean log-likelihood of their structural tokens and the highest is
scored, using no ground truth. Table~\ref{tab:protocol} re-scores one set of generations under three
alternatives, bounding what that choice costs. A best-of-$N$ oracle, the
reading that would make our tables incomparable to the baselines, is worth
$+1.4$ and $+1.3$ J\&F, so the reported rows lie within a point and a half of
any oracle reading and the multi-event outputs (2.2 per video) are not a
lottery the metric draws a winner from. Teacher-forcing the ground-truth expression, which every baseline in
Table~\ref{tab:grounding} is given for free, is worth about three times as much
($+4.2$, $+4.0$). Choosing \emph{which} event to report
is cheap; knowing \emph{what} to look for is what separates a query-free model
from a query-based one. Degenerate pointers scored the same way reach
$3.0$--$11.4$ J\&F, so the reported rows sit $40$ points above what the metric
returns without a working one.
\tabProtocol

\subsection{Spatiotemporal Grounding}
\label{sec:grounding}

Can a \emph{query-free} model localize as well as methods handed the answer in
text? Ref-YouTube-VOS is the stress test: every baseline receives a
ground-truth expression at test time. ReferFormer~\cite{wu2022referformer},
UNINEXT~\cite{yan2023uninext} and OnlineRefer~\cite{wu2023onlinerefer} pair
larger Swin-L or ViT-H backbones with an explicit query, while \tcam\ uses
ViT-B/16 and recovers masks by projecting its pointer weights onto the grid.
Table~\ref{tab:grounding} shows \tcam\ at 60.5 J\&F, within two to three points
of ReferFormer (62.9) and OnlineRefer (63.5) and trailing the much larger
UNINEXT (70.1), despite never seeing the expression every other row is handed
and never training on this dataset. Figure~\ref{fig:qual} (top) shows the
multi-subject case: two players active in one frame, each given an independent
trajectory set and caption with no query naming either.
\figQual
\tabGrounding

\subsection{Dense Video Captioning}
\label{sec:captioning}

Table~\ref{tab:captioning} evaluates the language quality of generated event
descriptions on ActivityNet Captions~\cite{krishna2017dense}. SODA\_c jointly
scores temporal localization and caption quality and is the primary metric.
PDVC~\cite{wang2021pdvc}, Vid2Seq~\cite{yang2023vid2seq} and Streaming
Vid2Seq~\cite{zhou2024streaming} are purpose-built dense captioners with
temporal localization heads, all trained on the ActivityNet Captions training
split that \tcam\ never sees. \tcam\ emits temporally grounded captions through
the format of Eq.~\ref{eq:output}, with time tokens converted to absolute
timestamps for metric computation. Transferred zero-shot, it improves over the
strongest in-domain baseline on all three metrics (6.5 vs.\ 6.2 SODA\_c, 39.4
vs.\ 37.8 CIDEr, 10.6 vs.\ 10.0 METEOR). The multi-event case at the bottom of
Figure~\ref{fig:qual}, where the sub-actions of one subject are split into
successive captioned intervals, is exactly the structure that a single
clip-level caption collapses into one sentence.

\tabCaptioning

\subsection{Trajectory Pointer Accuracy}
\label{sec:tapvid}

The \texttt{<TRK>} token activates the trajectory pointer
$\hat{\mathbf{p}}_i$ over the $K$ tracks. Is that pointer a genuine spatial
localizer or a soft attention map? We answer twice, because the two available
comparisons measure different things.

\noindent\textbf{Against chance, on the same backbone.} The pointer alone is responsible for selecting the \emph{right} tracks from
those its backbone supplies. We measure the fraction of selected tracks landing
within $\delta$ of an unoccluded ground-truth point, against the chance rate of
drawing as many uniformly from the same 1296. Only $4.0\%$ of the grid is on
target, yet across all 30 TAP-Vid DAVIS videos the pointer selects on-target
tracks at $5.3$--$8.5\times$ that rate (supp.\ Tab.~\ref{tab:pointerfid}),
falling monotonically in $k$ as the selected set is forced wider, so no $k$ is
load-bearing; thresholding the posterior at $\tau = 0.005$ drops $k$ entirely,
at $.294$ against $.306$ for fixed $k{=}32$ (supp.\ Tab.~\ref{tab:threshold}).

\noindent\textbf{Against dedicated trackers.} We also place the pointer beside
dedicated trackers under the TAP-Vid protocol (supp.\ Tab.~\ref{tab:tracking}). That comparison
is asymmetric and we state it plainly: TAPIR, PIPs++, CoTracker3 and
TAPNext~\cite{zholus2025tapnext} are scored on all query points, whereas \tcam\
is scored on the subset falling on the target its pointer selected, with the
top-$k$ activated tracks ($k{=}32$) as the prediction, so \tcam's row is not a
tracking result and should not be read as one. The selected trajectories come from CoTracker3 itself. Its full-query AJ is 64.5 on DAVIS and 54.4 on Kinetics, while the selected-subset readout scores 64.8 and 53.9. These describe different query populations and do not establish tracking parity or bound the cost of selection. The like-for-like pointer-selection test above is the primary evidence for grounding; \S\ref{sec:robust} tests transfer to another tracker. Supp.\ Fig.~\ref{fig:trackqual} shows
the pointer holding one subject through occlusion in a cluttered retail aisle.

\subsection{Counting and Sub-Object Attribution}
\label{sec:counting}

The two claims separating \tcam\ from an object-level captioner are that it
works out \emph{how many} events and subjects a video contains, and that it
grounds them below object granularity. Both are measured, not illustrated.

\noindent\textbf{Are the counts right?} Table~\ref{tab:countpmr}(a) scores the predicted count against annotation,
referenced to always-predict-one, which attains perfect precision and is what a
single-event model silently reproduces. On STGR dense captions
the event count reaches F1 $.690$ against $.405$ (MAE $2.94 \to 1.92$,
exact-match up five-fold). For subjects we cluster events by pointer cosine
similarity at the same $\delta = 0.3$ that $\mathcal{L}_{div}$ uses, adding no
hyperparameter, and compare against distinct annotated objects per MeViS video:
F1 $.723$ against $.483$. Recall trails precision for both ($.54$, $.61$), indicating that under-segmentation is the larger counting error.

\noindent\textbf{Is the grounding really sub-object?}
Mask supervision marks every track inside an object as positive, so the
objection that the pointer is only an object selector is the right one to
raise. We test it with the \emph{part-mass ratio}: pointer mass on a moving
part of an object over mass on the whole object, normalized by their area
ratio. The null is not arbitrary: a mask-uniform pointer scores exactly $1.0$,
providing a reference for uniform attribution within the object. Values above $1.0$ indicate greater concentration on the part; this reference is not an upper bound on all mask-supervised models. Pairs come from SA-V manual masklets~\cite{ravi2024sam2} by containment
and area ratio, the moving split further requiring the part to out-displace its
object.

With mask-BCE training, \tcam\ places $1.51\times$ the null share of its mass on
the moving part, $79\%$ of pairs above null (Table~\ref{tab:countpmr}(b));
replacing the flat on-object target with the motion-weighted one of
\S\ref{sec:loss} lifts this to $1.69\times$ on the moving split, $87\%$ above.
The comparison is paired and matched, each motion-weighted run against a
uniform-target control at the same loss weight, so the gain follows from the
weighting rather than the added term, and survives Bonferroni correction
($p = .0003$); supp.\ Tab.~\ref{tab:pmrmatched} replicates it over five matched finetunes, where
reweighting a \emph{uniform} target is flatly null ($p = .782$).
Figure~\ref{fig:subobject} is the qualitative counterpart. The result supports motion-weighted targets for finer attribution with the same masks.

\figSubObject
\tabCountPMR

\subsection{Robustness, Scale, and Backbone Dependence}
\label{sec:robust}

\noindent\textbf{Camera motion and visual disruption.} Using \S\ref{sec:formulation} we split MeViS Val into tertiles by global camera
displacement. Grounding does not degrade as the camera moves; it improves, by
$+5.3$ J\&F from the low to the high tertile ($r = +0.23$), because residual
motion separates a moving subject from a moving background more sharply than
from a static one, and explicit camera compensation changes nothing ($+0.2$
J\&F, matched; supp.\ Tab.~\ref{tab:camera}). The egocentric case is the extreme form of this test: transferred
zero-shot to EPIC-KITCHENS VISOR~\cite{darkhalil2022visor}, where the camera is
head-mounted and $84\%$ of tracks move, \tcam\ matches its in-domain MeViS J\&F
to within $0.3$ points. Scene cuts we measured rather than assumed: half the training videos already
contain one, VISOR is $3.8\times$ more disrupted than MeViS Val, and grounding
is unchanged across that range (supp.\ Tab.~\ref{tab:disruption}), because the pointer reads a
distribution over 1296 tracks rather than one, absorbing identity loss on
some.

\noindent\textbf{Track density and backbone dependence.} Since \car\ compresses any number of tracks into the same $M = 64$ tokens, $K$
never reaches the language model. Quadrupling $K$ from 576 to 2304 costs
$+60\%$ peak memory rather than $4\times$, adds no latency, and moves J\&F by
$0.9$ points (supp.\ Tab.~\ref{tab:scalability}): $36^2$ is a plateau, not a tuned point. Small fast objects a sparse grid under-samples, such as the ball missed in
Fig.~\ref{fig:qual}, are not recovered by denser sampling: the limit is
residual motion, not resolution. A carried ball moves with its holder and is
absorbed into that subject; once it acquires motion of its own, as at the
bounce in Fig.~\ref{fig:subobject}, the pointer separates it.
Finally, swapping the frozen backbone for
BootsTAPIR~\cite{doersch2024bootstap}, an unrelated family built on cost-volume
matching rather than joint transformer attention, changes pointer-selection
accuracy by at most $0.021$ at any $k$ with the sign flipping, \emph{without
retraining}; a matched retraining moves J\&F by $0.2$ points and improves
event-count F1 by $.005$ (supp.\ Tab.~\ref{tab:trackerswap}). \tcam\ is a tracker-agnostic
language interface to point tracking.

\subsection{Ablation Studies}
\label{sec:ablations}

\noindent\textbf{Component ablation.}
Supp.\ Tab.~\ref{tab:ablation} removes each module and training term in turn. Dropping
$\mathcal{L}_{grd}$ causes the largest collapse on both grounding metrics (J\&F
$52.0\!\to\!27.8$, AJ $64.8\!\to\!24.5$) while leaving METEOR almost untouched
($10.6\!\to\!10.4$): caption fluency and grounding are learned largely
independently, and the pointer has no incentive to stay on-object without it. Removing the Trajectory Tokenizer (mean-pooled ViT features) deletes
point-level motion outright, and AJ becomes unmeasurable; removing the
VisionMotionEncoder costs every metric a smaller but consistent amount.
Removing CAR degrades caption quality most, as the $T \times K$ sequence
overwhelms T5's attention, and removing $\mathcal{L}_{div}$ leaves captions
intact but collapses J\&F.

\noindent\textbf{Structural token supervision and data scaling.}
\label{sec:struct_weight_scale}
Supp.\ Tab.~\ref{tab:struct_weight} reports the structural upweight $w_{struct}$, which controls
generation of the structural tokens.
Trained from scratch without upweighting the model fails to terminate events
reliably and reverts to single-event outputs (12.4\% multi-event rate); at
$w_{struct} = 1.5$ that becomes 86.5\% with the best J\&F, while 3.0 and 5.0
buy marginal rate as J\&F falls to 49.1 and 46.3. This is a claim
about \emph{acquiring} the structural vocabulary, not a narrow optimum: once it
is acquired, a matched sweep over $[1.0, 3.0]$ moves J\&F by $0.5$ points and
event-count F1 by $.015$ (supp.\ Tab.~\ref{tab:structplateau}), so 1.5 is a plateau we adopt rather
than a peak we tuned to. Growing the training set from MeViS alone (44.8 J\&F, 7.2 METEOR) through
$+$STGR (48.7, 8.9) to full PLM-Video-Human (52.0, 10.6) improves both metrics
with scale (supp.\ Tab.~\ref{tab:data_scale}).
\section{Conclusion}
\label{sec:conclusion}
We presented \tcam, which generates captions, intervals, and trajectory pointers for multiple motion events without a query or region prompt. A Caption-Aware Resampler connects dense tracks to a language decoder, supporting zero-shot transfer to ActivityNet Captions and Ref-YouTube-VOS. Confidence-based event selection, count evaluation, and sub-object attribution diagnostics complement the benchmark comparisons. The tested camera-motion and disruption subsets show similar or improved grounding, but do not establish robustness to all scene cuts. Remaining limits include binary mask supervision, a fixed 16-frame time vocabulary, and under-segmentation in busy scenes. Future work includes graded motion supervision, learned window linking, and 3D trajectories~\cite{xiao2024spatialtracker,zhang2025tapip3d}.

{
  \small
  \bibliographystyle{ieeenat_fullname}
  \bibliography{main}
}

\clearpage
\section*{Supplementary Material}
\renewcommand{\theHsection}{supp.\arabic{section}}
\renewcommand{\theHtable}{supp.\arabic{table}}
\renewcommand{\theHfigure}{supp.\arabic{figure}}

\renewcommand{\theHsection}{supp.\arabic{section}}
\renewcommand{\theHtable}{supp.\arabic{table}}
\renewcommand{\theHfigure}{supp.\arabic{figure}}

\renewcommand{\thetable}{S\arabic{table}}
\renewcommand{\thefigure}{S\arabic{figure}}
\setcounter{table}{0}
\setcounter{figure}{0}
\setcounter{section}{0}
\renewcommand{\thesection}{S\arabic{section}}
\renewcommand{\theHsection}{supp.\arabic{section}}
\renewcommand{\theHtable}{supp.\arabic{table}}
\renewcommand{\theHfigure}{supp.\arabic{figure}}

This document reports the experiments referenced from the main paper. Unless a
matched finetune is stated, every row uses the same released checkpoint as the
main paper, the same query-free selection protocol (confidence top-1,
\S4.1 of the main paper), and the same 50-video MeViS \texttt{valid\_u} subset
with public masks. \textbf{Camera-motion, disruption, track-density, tracker-swap, and matched structural-weight results report J\&F differences against their stated reference rows}, because these ablations were run under a
single reduced-schedule protocol in which only the contrast between matched
arms is meaningful.

\section{Camera Motion and Visual Disruption}
\label{sec:supp_robust}

\noindent\textbf{Defining ``moving''.}
The main paper (\S3.1) defines a track as moving when its \emph{residual} speed,
after a global affine transform $\mathbf{H}_t$ fitted by RANSAC over all
$K = 1296$ tracks is removed, exceeds
$\text{med}(\text{inliers}) + 2\,\text{MAD}(\text{inliers})$. The background is
the dominant inlier set, so this separates object motion from camera motion
without any annotation, and the resulting per-video camera displacement is what
we stratify by.

\noindent\textbf{Stratification.}
Table~\ref{tab:camera} splits MeViS Val into camera-motion tertiles. Grounding
does not degrade as the camera moves; it improves by $+5.3$ J\&F from the low
to the high tertile, with Pearson $r = +0.23$ between camera displacement and
J\&F. The mechanism is visible in the last column: the fraction of tracks
classified as moving rises from $.25$ to $.57$ across the tertiles, because
residual motion discriminates a moving subject from a moving background more
sharply than from a static one, where a stationary subject and a stationary
background are equally residual-free.

\noindent\textbf{Testing an explicit fix.}
We also tested whether an explicit camera-compensation term helps, adding a
fourth branch to Eq.~2 of the main paper:
\begin{equation}
  \tau_{j,t} = \phi_{j,t} + \text{MLP}_{res}(\dot{\mathbf{x}}^{res}_{j,t}),
  \label{eq:tau}
\end{equation}
where $\dot{\mathbf{x}}^{res}$ is the residual displacement defined above. Two
matched 1.5K-step runs differ only in whether this branch is present. The mean
effect is $+0.2$ J\&F, neutral within noise (Table~\ref{tab:camera}, last row),
so we do not adopt it: the trajectory tokens already carry what a compensation
module would supply.

\tabCamera

\noindent\textbf{Egocentric transfer.}
The hardest camera-motion case is a camera worn on the subject's head.
Table~\ref{tab:disruption} evaluates all 43 EPIC-KITCHENS VISOR \emph{val}
videos zero-shot: $84\%$ of tracks are classified as moving, against $49\%$ on
MeViS Val, and grounding is unchanged to within $0.3$ J\&F points. Nothing in
the model was adapted for the egocentric domain.

\noindent\textbf{Scene cuts.}
The submitted paper claimed that scene cuts degrade pointer quality. We
measured this instead of assuming it. Detected shot boundaries appear in
$17.5\%$ of MeViS Val videos and, notably, in $50\%$ of the STGR/SA-V training
videos, so the model has already been trained through them. VISOR carries
$3.8\times$ the frame-to-frame appearance disruption of MeViS Val at the same
grounding quality. The pointer scores a distribution over all $K$ tracks rather
than following one track, so identity loss on a subset of tracks is absorbed
rather than propagated. We have withdrawn the claim from the main paper.

\tabDisruption

\section{Caption-Aware Resampler Scalability}
\label{sec:supp_scale}

Table~\ref{tab:scalability} varies only the CoTracker3 grid density on the same
videos under the same protocol. Three observations follow.

Accuracy is flat: J\&F spans $0.9$ points across a $4\times$ change in $K$
($576 \to 2304$), so the $36^2$ default is anywhere on a plateau rather than a
tuned operating point. Memory scales sub-linearly, $+60\%$ for $4\times$ the
tracks, and latency does not grow at all (the ordering in the table is warm-up
noise), because per-video cost is dominated by autoregressive beam search
rather than by the trajectory encoder. The reason is structural: the trajectory
path is $O(M \cdot T \cdot K \cdot d)$, linear in $K$, but \car\ emits a
\emph{fixed} $M = 64$ context vectors, so the T5 encoder-decoder context length is independent of $K$. The trajectory encoder and the pointer readout still process track-dependent inputs.

\tabScalability

\noindent\textbf{On longer temporal contexts.}
The structural time vocabulary $\texttt{<T}_0\texttt{>}\ldots\texttt{<T}_{15}\texttt{>}$
is fixed at $T = 16$ by the tokenizer, so $T$ cannot be varied without
retraining. \tcam's answer to longer video is to window rather than to extend
context: the per-video cost measured above is the per-window unit cost, and
window linking is linear in video length. This is a genuine limitation and is
stated as such in the conclusion.

\section{Backbone Dependence}
\label{sec:supp_tracker}

\tcam\ reads from a frozen tracker, so the natural question is how much of its
behaviour is CoTracker3's. We replace the backbone with
BootsTAPIR~\cite{doersch2024bootstap}, chosen because it belongs to a different
family: per-query cost-volume matching with iterative refinement, rather than
joint transformer attention across tracks.

\noindent\textbf{Without retraining.} The same checkpoint, pointer, protocol,
grid, and 30 TAP-Vid DAVIS videos; only the trajectories change.
Table~\ref{tab:trackerswap}(a) shows $|\Delta| \leq 0.021$ at every $k$, with
the sign flipping between $k$ values, which is noise-scale. The gain over each
backbone's own on-target coverage is preserved as well ($5.3$--$8.5\times$ for
CoTracker3, $4.6$--$7.3\times$ for BootsTAPIR). A pointer trained on
CoTracker3 trajectories transfers to BootsTAPIR trajectories out of the box.

\noindent\textbf{With retraining.} Two finetunes identical in every respect
except which tracker produced the cached trajectories
(Table~\ref{tab:trackerswap}(b)) differ by $0.2$ J\&F points, and event-count
F1 is marginally \emph{better} with BootsTAPIR. \tcam\ trains equally well on
either, so it is best understood as a tracker-agnostic language interface to
point tracking, which inherits improvements in tracking rather than being
bounded by one tracker.

\tabTrackerSwap

\section{Component Ablation}
\label{sec:supp_abl}

Table~\ref{tab:ablation} is the component ablation discussed in \S4.7 of the
main paper, whose numbers are quoted there in full.

\section{Structural Upweight is a Plateau}
\label{sec:supp_struct}

Table~\ref{tab:struct_weight} samples $w_{struct} \in \{1.0, 1.5, 3.0, 5.0\}$
across independently trained models and shows a large J\&F swing, which makes
the hyperparameter look brittle. Table~\ref{tab:structplateau} separates two
effects that this conflates. Table~\ref{tab:data_scale} reports data scaling.

Under a matched protocol, six 600-step finetunes from the released checkpoint
identical except for $w_{struct}$, J\&F varies by $0.5$ points and event-count
F1 by $.015$ across the whole span $[1.0, 3.0]$, including at the three points
Table~\ref{tab:struct_weight} samples. Both response surfaces are flat, and the two axes move in
opposite directions: J\&F rises slightly with $w_{struct}$ while event-count F1
falls slightly, which is the expected trade-off between structural emphasis and
mask fidelity and is what makes $1.5$ a reasonable default rather than a tuned
peak.

What Table~\ref{tab:struct_weight} measures is therefore not sensitivity around the operating point
but whether the structural vocabulary is \emph{acquired at all}. A model
trained from scratch at $w_{struct} = 1.0$ never learns to emit
\texttt{<CAP\_END>} reliably and collapses to single-event outputs; a short
finetune from a model that already emits it cannot undo that, which is why
every row here retains the multi-event advantage (event-count F1
$.675$--$.690$ against the always-predict-one baseline of $.405$).

\tabAblation
\tabStructWeight
\tabDataScale
\tabStructPlateau

\section{Sub-Object Attribution at a Real Budget}
\label{sec:supp_pmr}

\S4.5 of the main paper reports the part-mass ratio for the released checkpoint
and for a short motion-weighted finetune. Table~\ref{tab:pmrmatched} is the
controlled replication at a larger budget: five matched 3K-step finetunes on
the moving-part split, each motion-weighted arm paired with a uniform-target
control at the \emph{same} $\lambda_{sm}$.

The design is what makes the claim causal. If the improvement came from adding
a softmax competition term at all, then raising $\lambda_{sm}$ from $1.0$ to
$4.0$ with a \emph{uniform} target would move the metric; it does not
($p = .782$ on the moving split, $p = .814$ on containment). Only the
motion-weighted target moves it, at every $\lambda_{sm}$ tested, and all three
motion-weighted comparisons clear the Bonferroni threshold of
$.05/8 = .00625$ over the eight tests run.

Two sanity checks support the metric itself. Replacing the pointer with a
uniform distribution reproduces the null exactly across all curated pairs
(maximum deviation $2.2 \times 10^{-16}$), confirming that a mask-uniform
pointer scores $1.0$ by construction. And the positive \emph{set} is unchanged
by the motion weighting (verified at 509, 402 and 283 on-object tracks on
sampled examples): the binary term still says ``point somewhere on the
object'', and only the competition among on-object tracks is graded.

\tabPMRmatched

\section{Pointer Fidelity Without a Fixed \texorpdfstring{$k$}{k}}
\label{sec:supp_tau}

This section supports the two claims \S4.4 of the main paper makes about the
pointer, both of which compare it against the \emph{same} frozen backbone it
reads from, so that the readout is isolated from the tracker.

\noindent\textbf{The pointer beats chance at every $k$.}
Backbone coverage on TAP-Vid DAVIS is $0.0398$: only $4.0\%$ of the
$1296$-track grid lies on the annotated target. That is the chance rate of
drawing $k$ tracks uniformly, and the rate the pointer would reach in the limit
$k \to K$, so it is the correct null for a selection metric and the reason
these ratios are not an artifact of large targets.
Table~\ref{tab:pointerfid} reports selection accuracy against it across the
whole range of $k$. The ratio falls monotonically, from $8.5\times$ at $k = 8$
to $5.3\times$ at $k = 128$, as the selected set is forced wider than the
target: no single $k$ is load-bearing. This is the like-with-like comparison,
scored on all 30 videos under one protocol, and it is the measure that carries
the pointer claim; the tracker table of \S\ref{sec:supp_tapvid} does not share
its protocol and does not.

\noindent\textbf{The readout does not need $k$ at all.}
Because the choice of $k$ is an evaluator's choice, we also remove it. Taking
the predicted set as $\{k : \sigma(s_{i,k}) > \tau\}$ lets the model choose how
many tracks to select per event (Table~\ref{tab:threshold}). At
$\tau = 0.005$ the mean selected set is $55.2$ tracks at accuracy $.294$
against $.306$ for fixed $k = 32$, so every number in \S4.4 can be restated
with no reference to $k$.

\tabPointerFidelity
\tabThreshold

\section{TAP-Vid Comparison Table}
\label{sec:supp_tapvid}

Table~\ref{tab:tracking} is the tracker comparison discussed in \S4.4 of the
main paper, reproduced here because its protocol is not shared with the other
rows. The dedicated trackers are scored on all TAP-Vid query points; \tcam\ is
scored on the subset falling on the target its pointer selected, taking the top
$k = 32$ activated tracks as the prediction. \tcam's row is therefore not a
tracking result. The selected trajectories are supplied by CoTracker3, but the different query populations prevent interpreting the numerical gap as a bound on readout cost. Pointer selection is evaluated directly in \S\ref{sec:supp_tau}.

\tabTracking

\section{Additional Figures}
\label{sec:supp_figs}

Figure~\ref{fig:resampler} details the Caption-Aware Resampler, and
Figure~\ref{fig:trackqual} shows the trajectory pointer qualitatively in a
cluttered retail environment. Both were moved here from the main paper to make
room for the quantitative results above.

\figResampler
\figtrackingQual

\end{document}